\documentclass[runningheads,review]{llncs}
\usepackage{amsmath, amsxtra, amsfonts, amssymb, amstext}
\usepackage{lineno}
\usepackage{nicefrac}
\usepackage{xspace}
\usepackage{graphics,color}
\usepackage[colorlinks]{hyperref}
\definecolor{linkblue}{rgb}{0.1,0.1,0.8}
\hypersetup{colorlinks=true,linkcolor=linkblue,filecolor=linkblue,urlcolor=linkblue,citecolor=linkblue}
\usepackage{float}
\usepackage{etoolbox}

\usepackage[ruled,vlined,linesnumbered]{algorithm2e}

\usepackage{footnote}
\usepackage{enumerate}

\usepackage{tikz}
\usetikzlibrary{trees,positioning,shapes,arrows}
\tikzstyle{line}=[draw]
\tikzstyle{level 1}=[level distance=3.5cm, sibling distance=6cm]
\tikzstyle{level 2}=[level distance=5cm, sibling distance=4cm]
\tikzstyle{level 3}=[level distance=7cm, sibling distance=6cm]
\tikzstyle{level 4}=[level distance=7cm, sibling distance=4cm]

\tikzstyle{bag} = [text width=4em, text centered]
\tikzstyle{end} = [circle, minimum width=3pt,fill, inner sep=0pt]

\newcommand{\R}{\mathbb{R}}

\newcommand{\E}{\mathbb E}
\renewcommand{\epsilon}{\varepsilon}
\newcommand{\eps}{\varepsilon}

\newcommand{\ooea}{\ensuremath{(1 + 1)}\text{-EA}\xspace}
\newcommand{\toea}{\ensuremath{(2 + 1)}\text{-EA}\xspace}
\newcommand{\thoea}{\ensuremath{(3 + 1)}\text{-EA}\xspace}
\newcommand{\fiveoea}{\ensuremath{(5 + 1)}\text{-EA}\xspace}
\newcommand{\toga}{\ensuremath{(2 + 1)}\text{-GA}\xspace}
\newcommand{\thoga}{\ensuremath{(3 + 1)}\text{-GA}\xspace}
\newcommand{\fiveoga}{\ensuremath{(5 + 1)}\text{-GA}\xspace}
\newcommand{\olea}{\ensuremath{(1 + \lambda)}\text{-EA}\xspace}
\newcommand{\moea}{\ensuremath{(\mu + 1)}\text{-EA}\xspace}
\newcommand{\moga}{\ensuremath{(\mu + 1)}\text{-GA}\xspace}
\newcommand{\ollga}{\ensuremath{(1 + (\lambda,\lambda))}\text{-GA}\xspace}

\newcommand{\mogan}{\ensuremath{(\mu + 1)}\text{-GA-NoCopy}\xspace}
\newcommand{\togan}{\ensuremath{(2 + 1)}\text{-GA-NoCopy}\xspace}

\newcommand{\onemax}{\textsc{OneMax}\xspace}

\newcommand{\dynbv}{\textsc{DynBV}\xspace}

\newtoggle{journal}
\toggletrue{journal}
\begin{document}

\title{Large Population Sizes and Crossover Help in Dynamic Environments}
%\titlenote{Produces the permission block, and copyright information}
%\subtitle{Extended Abstract}
%\subtitlenote{The paper contains only rough proof sketches. We have attached all missing proof in an appendix, in case a reviewer wants to look up any further details. The appendix will be removed for the final version.}

%\author{Johannes Lengler}
%%\authornote{Dr.~Trovato insisted his name be first.}
%%\orcid{1234-5678-9012}
%\affiliation{%
%  \institution{ETH Z{\"u}rich, Switzerland}
%%  \streetaddress{P.O. Box 1212}
%%  \city{Dublin} 
%% \state{Ohio} 
%% \postcode{43017-6221}
%}
\author{Johannes Lengler, Jonas Meier}
\authorrunning{J. Lengler, J. Meier}
\institute{Department of Computer Science \\ETH Z{\"u}rich, Switzerland}
\maketitle
\begin{abstract}
Dynamic linear functions on the hypercube are functions which assign to each bit a positive weight, but the weights change over time. Throughout optimization, these functions maintain the same global optimum, and never have defecting local optima. Nevertheless, it was recently shown [Lengler, Schaller, FOCI 2019] that the $(1+1)$-Evolutionary Algorithm needs exponential time to find or approximate the optimum for some algorithm configurations. In this paper, we study the effect of larger population sizes for Dynamic BinVal, the extremal form of dynamic linear functions. We find that moderately increased population sizes extend the range of efficient algorithm configurations, and that crossover boosts this positive effect substantially. Remarkably, similar to the static setting of monotone functions in [Lengler, Zou, FOGA 2019], the hardest region of optimization for $(\mu+1)$-EA is not close the optimum, but far away from it. In contrast, for the $(\mu+1)$-GA, the region around the optimum is the hardest region in all studied cases.
\iftoggle{journal}{}
{% if journal = false
\footnotetext{Extended Abstract. A full version is available on arxiv at~\cite{fullversion}.}
}
\end{abstract}

% \begin{CCSXML}
%<ccs2012>
%<concept>
%<concept_id>10003752.10010070.10011796</concept_id>
%<concept_desc>Theory of computation~Theory of randomized search heuristics</concept_desc>
%<concept_significance>500</concept_significance>
%</concept>
%</ccs2012>
%\end{CCSXML}
%
%\ccsdesc[500]{Theory of computation~Theory of randomized search heuristics}
%
%%\category{F.2.2}{Theory of Computation}{Analysis of Algorithms and Problem Complexity}[Nonnumerical Algorithms and Problems]
%\keywords{Theory, Runtime Analysis, Monotone Functions, Crossover, Mutation Strength, HotTopic}
%%\textbf{Category:} {F.2.2}{Theory of Computation}{Analysis of Algorithms and Problem Complexity}[Nonnumerical Algorithms and Problems]\\
%%
%%\textbf{Keywords:} {Black-Box Complexity, Elitist Selection, Comparison-Based Algorithms}
%

%\newcommand{\package}{\emph}

\section{Introduction}\label{sec:intro}

The $(\mu+1)$ Evolutionary Algorithm and the $(\mu+1)$ Genetic Algorithm, \moea and \moga for short, are heuristic algorithms that aim to optimize an objective or fitness function $f : \{0,1\}^n \rightarrow \mathbb{R}$. Both maintain a population of $\mu$ search points, and in each round they create an offspring from the population and discard one of the $\mu+1$ search points, based on their objective values. They differ in how the offspring is created: the \moea chooses a parent from the population and \emph{mutates} it, the \moga uses \emph{crossover} of two parent solutions in addition to mutation.

Two classical theoretical questions for these algorithms have ever been: 
\begin{itemize}
\item What is the effect of the population size? In which optimization landscapes and regimes are larger (smaller) populations beneficial?
\item In which situations does crossover improve performance? 
\end{itemize}
Although these questions have ever been central for studies of the \moea and the \moga, there is still vivid ongoing research on these questions, see~\cite{antipov2018tight,dang2017escaping,doerr2012crossover,jansen2002analysis,pinto2018simple,sudholt2017crossover,witt2006runtime,witt2013tight} for a selection of theoretical work. (Also, the book chapter~\cite{sudholt2020benefits} treats related topics.) More generally, the research question is: which algorithm configurations perform well in which optimization landscapes? Such landscapes are given by a specific benchmark functions or by a class of functions.

Recently, a new type of dynamic landscapes was introduced by Lengler and Schaller~\cite{ulyjohan2018}. It was called \emph{noisy linear functions} in~\cite{ulyjohan2018}, but we prefer the term~\emph{dynamic linear functions}. In this setting, the objective function is of the form $f:\{0,1\}^n \rightarrow \mathbb{R}; f(x) = \sum_{i=1}^n W_ix_i$ with positive coefficients $W_i>0$. However, the twist is that the weights $W_i$ are redrawn for each generation. I.e., we have a distribution $\mathcal D$, and for the $t$-th generation we draw i.i.d.~weights $W_i^{(t)}$ from that distribution, which define a function $f^{(t)}$. Then the $\mu+1$ competing individuals are compared with respect to the fitness function $f^{(t)}$. 

To motivate this setting, let us give a grotesquely oversimplified example. Imagine a chess engine has $100$ bits as parameters. Each bit switches on/off a database tailored to one specific opening ($1 =$ access, $0 =$ no access), and this improves performance massively for this opening. E.g., the first bit determines whether the engine plays well in a French Opening, but has no influence whatsoever on the performance of an Italian Opening (the database is ignored since it does not produce matches to that situation). Let us go even further and assume that the engine will always win an opening if the corresponding database is active, and always lose with inactive database. Then we have removed even the slightest ambiguity, and this setting has a obvious optimal solution, which is the all-one string (activate all databases). This situation may seem completely trivial, but crucially, \emph{it is not solved by some standard optimization algorithms}. 

To complete the analogy, assume that the engine is trained by playing against different players, where player $t$ has probability $W_i^{(t)}$ to choose the $i$-th opening. Then the reward is precisely the dynamic linear function introduced above, and it was shown in~\cite{ulyjohan2018} that \emph{the \ooea needs exponential time to approximate the optimum within a constant factor} when configured with bad parameters. These bad parameter settings look quite innocent. With standard bit mutation (i.e., for mutation we flip each bit independently with probability $p=c/n$), any choice $c > c_0 \approx 1.59$ leads asymptotically to an exponential time for finding or approximating the optimum, if the distribution $\mathcal D$ is too skewed. On the other hand, for any $c<c_0$ the \ooea finds the optimum in time $O(n\log n)$ for any~$\mathcal D$. This lack of stability motivates our paper: we ask whether larger population sizes and/or crossover can push the threshold $c_0$ of failure.

Optimization of dynamic functions may occur in various contexts. The chess engine with varying opponents is one such example. Similar examples arise in the context of co-evolution, e.g., a chess-engine trained against itself, or a team of DOTA agents in which some abilities of an agent (good aim, good exploration strategy, good path planning, ...) are always helpful (positive weight), but may be more or less important depending on her current co-agents. A rather different example is planning the timetable of a transportation company: to be efficient in the exploration phase of the optimization algorithm, schedules may be compared only for some partial data, and not for the whole data set. Similarly, consider an optimization process in which the function evaluation involves an offline test, as in drug development or robotic training. Then each test may involve subtly varying outer conditions (e.g., different temperatures or humidity, different lighting), which effectively gives a slightly different fitness function for each test.

 \noindent \textbf{Our Results in a Nutshell.}
Instead of the full range of dynamic linear functions as in~\cite{ulyjohan2018}, we only study the limiting case of these functions, which we call~\emph{dynamic binval}. We perform experiments to study the performance of the \moea and the \moga for small values of $\mu$. Similarly as for the \ooea, we find that for each algorithm there is a threshold $c_0$ such that the algorithm is efficient for every mutation parameter $c<c_0$, and inefficient for $c>c_0$. This threshold~$c_0$ is our main object of study, and we investigate how it depends on the algorithmic choices. We find that an increased population size helps to push~$c_0$, but that the benefits are much larger when crossover is used. As a baseline, we recover the theoretical result from~\cite{ulyjohan2018} that for the \ooea the threshold is at $c_0 \approx 1.59$, though experimentally, for $n=3000$ it seems closer to $1.7$. For the \toea the threshold increases to $c_0\approx 2.2$, and further to $c_0 \approx 3.1$ for the \toga. If we explicitly forbid that the two parents in crossover are identical then the threshold even shifts to $c_0 \approx 4.2$. We call the resulting algorithm \togan. For larger population sizes we get a threshold of $c_0 \approx 2.6$ for the \thoea, $c_0 \approx 3.4$ for the \fiveoea, $c_0 \approx 6.1$ for the \thoga, and $c_0 > 20$ for the \fiveoga.

The theoretical results for the \ooea predict that the runtime jumps from quasi-linear to exponential. Indeed, we can experimentally confirm huge jumps in the runtime even for slight changes of the mutations parameter $c$. For example, we obtain a significant $p$-value for the a posteriori hypothesis that the \toea with $c=2.5$ is more than $60$ times slower than the \toea with $c=2.0$. In fact, this is a highly conservative estimate since we needed to cut off the runs for $c=2.5$. We systematically list these factors in our result sections.

To get a better understanding of the hardness of the optimization landscape, we compute the \emph{drift of degenerate populations}, inspired by~\cite{Dichotomy}. We call a population \emph{degenerate} if it consists entirely of multiple copies of the same individual. If $X_i$ is the number of zero-bits in the $i$-th degenerate population, then we estimate the drift $\E[X_{i}-X_{i+1} \mid X_i = y]$ by Monte-Carlo simulations. Moreover, for $y$ close to $0$ we derive precise asymptotic formulas for the degenerate population drift for the \toea and the \toga. In~\cite{Dichotomy} the degenerate population drift was studied theoretically for the \moea on monotone functions, which is a related, but not identical setup (see below). Still, part of the analysis carries over: if the population drift is negative for some $y$ then the runtime is exponential, while it is $O(n\log n)$ if the population drift is positive everywhere.
 
Perhaps surprisingly, the \moea and the \moga are not just quantitatively different, but we also find a strong qualitative difference in the hardness landscape. For the \moga, the ``hardest'' part of the optimization process is close to optimum, in all cases that we have experimentally explored. Formally, we found that if the degenerate population drift is negative \emph{somewhere}, then it is also negative close to the optimum. For the \moea, we found the opposite: the degenerate population drift can be negative for some intermediate ranges, although it is positive around the optimum. This implies that the hard part of optimization (taking exponential time) is getting in the vicinity of the optimum. But once the algorithm is somewhat near the optimum, it will efficiently finish optimization. This behavior is rather counter-intuitive, since common wisdom says that optimization gets harder close to the optimum. Notably, a similar phenomenon has recently been proven for certain monotone functions by Lengler and Zou~\cite{Dichotomy,ExpSlowdown}, see below.

\noindent \textbf{Related Work.} The only previous work on dynamic linear functions is by Lengler and Schaller~\cite{ulyjohan2018}. As mentioned before, they proved that for every $c>c_0\approx 1.59$ there is $\eps>0$ and a distribution $\mathcal D$ such that the \ooea with mutation rate $c/n$ needs exponential time to find a search point with at least $(1-\eps)n$ one-bits for dynamic linear functions with weight distribution $\mathcal D$. For $c<c_0$ the optimization time is $O(n\log n)$ for all distributions~$\mathcal D$. Moreover, for any $c>c_0$, they gave a completely characterization of all distributions for which the \ooea with mutation rate $c/n$ is efficient/inefficient.

An important strand of work that is similar in spirit, though not in detail, is the study of \emph{monotone functions}. A function is \emph{monotone} if for every bit-string, flipping any zero-bit into a one-bit increases the fitness. Doerr, Jansen, Sudholt, Winzen, and Zarges \cite{doerr2013} and Lengler and Steger \cite{lengler2018drift} showed that there are monotone functions for which the \ooea needs exponential time to find or approximate the optimum if the mutation parameter $c$ is too large ($c > c_0 \approx 2.1$), while it is efficient for all monotone functions if $c\leq 1+\eps$ for some small $\eps>0$~\cite{lengler2019does}. The construction of hard (static) instances from \cite{lengler2018drift} was named HotTopic in~\cite{Dichotomy}, and it resembles dynamic linear functions: the HotTopic function is locally given by linear functions with certain positive weights, but as the algorithm proceeds from one part of the search space (``level'') to the next, the weights change. This analogy inspired the introduction of dynamic linear functions in~\cite{ulyjohan2018}. 

For HotTopic functions, there is a plethora of results. In~\cite{Dichotomy}, the dichotomy between exponential and quasi-linear time from the \ooea was extended to a large number of other algorithms, including the \olea, the \moea, their so-called ``fast'' counterparts, and the \ollga. On the other hand, it was shown that the \moga is always efficient for HotTopic functions if the population size is sufficiently large, while for the \moea the population size does not change the threshold $c_0$ at all. Notably, for the population-based algorithms \moea and \moga, the efficiency result was only obtained for parameterizations of the HotTopic functions in which the weight changes occur close to the optimum. This seemed like a technical detail at first, but in an extremely surprising result, Lengler and Zou~\cite{ExpSlowdown} showed that this detail was hiding an unexpected core: if the weights are changed far away from the optimum, then increasing the population size has a devastating effect on the performance of the \moea. For any $c>0$ (also values much smaller than $1$), there is a $\mu_0$ such that the \moea with $\mu \geq \mu_0$ and mutation rate $c/n$ needs exponential time on some monotone functions. Together with~\cite{Dichotomy}, this shows three things for monotone functions:
\begin{enumerate}
\item For optimization close to the optimum, the population size has no strong impact on the performance of the \moea.
\item Close to the optimum, the \moga outperforms the \moea massively (quasi-linear instead of exponential) if the population size is large enough. It can cope with any constant mutation parameter $c$.
\item Far away from the optimum, a larger population size decreases the performance of \moea massively. There is no safe choice of $c$ if $\mu$ is too large.
\end{enumerate}
It would be extremely interesting to understand the \moga far away from the optimum. Unfortunately, such results are unknown. Theoretical analysis is hard (though perhaps not impossible), and function evaluations of HotTopic are extremely expensive, so experiments are only possible for very small problem sizes. Our paper can be seen as the first work which studies the behavior of the \moga in a related, though not identical setting.

We conclude this section with a word of caution. HotTopic functions and dynamic linear functions are similar in spirit, but not in actual detail. For example, the analysis of the \moga in~\cite{Dichotomy} or of the \moea in~\cite{ExpSlowdown} rely heavily on the fact that there weights are locally stable in HotTopic functions. Thus it is unclear how far the analogy carries. Some of our experimental findings for the \moea for dynamic linear functions differ from the theoretical (asymptotic) results for HotTopic in~\cite{Dichotomy,ExpSlowdown}. For us, a larger $\mu$ is beneficial, as it shifts the theshold $c_0$ to the right. For HotTopic functions, it does not shift the threshold at all if the algorithm operates close to the optimum, and it shifts the threshold \emph{to the left} (i.e., makes things worse) far away from the optimum. This could either be because the theoretical effects only kick in for very large $\mu$, or because HotTopic and dynamic linear functions are genuinely different. On the other hand, both settings agree in the surprising effect that the hardest part for the algorithm is not close to the optimum, but rather far away from it.

\section{Preliminaries}

\subsection{The Algorithms}

All our considered algorithms maintain a population $P$ of search points of size $\mu$. In each round (or \emph{generation}), they create an offspring from the population, and from the $\mu+1$ search points they remove the one with lowest fitness, breaking ties randomly. They only differ in the offspring creation. The \moea uses \emph{standard bit mutation}: a random parent is picked from the population, and each bit in the parent is flipped with probability $c/n$. The genetic algorithms flips a coin in each round whether to use mutation (as above), or whether to use \emph{bitwise uniform crossover}: for the latter, it picks two random parents from the population, and for each bit it randomly chooses the bit of either parent. For the \moga, the two parents are chosen independently. For the \mogan, they are chosen without repetition. The parameters are thus the mutation parameter $c>0$, which we will assume to be independent of $n$, and the population size $\mu$. In our theoretical (asymptotic) results, we will only consider $\mu=2$. The pseudocode description is given in Algorithm~\ref{alg:all}.

\begin{algorithm}
% \textbf{Initialization:} \\
% \Indp
Initialize $P$ with $\mu$ independent $x \in \{0,1\}^n$ uniformly at random\;
% \For{$i=1,\ldots,\mu$}{
%Sample $x^{(i)}$ uniformly at random from $\{0,1\}^n$\;
% $X \leftarrow X \cup \{ x^{(i)}\}$\;
% }% For i
% \Indm
 \textbf{Optimization:}	
 \For{$t=1,2,3,\ldots$}{
% \For{$i=1,2,\ldots \lambda$}{
\textbf{Creation of Offspring:} For GAs, flip a fair coin to do either mutation or crossover; for EA, always do mutation and no crossover. \\
                 \If{mutation}{
		Choose $x\in P$ uniformly at random\;
		\label{line:mutation} Create $y$ by flipping each bit in $x$ independently with probability $c/n$\;}
		\label{line:crossover}\If{crossover}{  
		Choose $x,x'\in X$ uniformly at random: independently for \moga; without repetition for \mogan\;
		Create $y$ by setting $y_i$ to either $x_i$ or $x_i'$, each with probability $1/2$, independently for all bits\; \label{line:endcrossover} }
	 Set $P \leftarrow P \cup \{y\}$\;
%	 }%For i
\textbf{Selection:}\label{line:selection} Select $z \in \arg\min \{f(x)\mid x\in P\}$ (break ties randomly) and update $P \leftarrow P \setminus \{z\}$\;
	 }%For t
\caption{{\moea, \moga, and \mogan with mutation parameter~$c$ for maximizing an unknown function $f:\{0,1\}^n \rightarrow \R$. $P$ is a multiset, i.e., it may contain search points several times.}}
\label{alg:all}
\end{algorithm}

\subsection{Dynamic Linear Functions and the Dynamic Binval Function}\label{sec:dynbinval}

We have described the algorithms for optimizing a static fitness function $f$. However, throughout the thesis, we will consider dynamic functions that changes in every round. We denote the function in the $t$-th iteration by $f^{(t)}$. That means that in the selection step (Line~\ref{line:selection}), we select the worst individual as $z \in \arg\min \{f^{(t)}(x)\mid x\in P^{(t)}\}$, where $P^{(t)}$ is the $t$-th population. Crucially, we never mix different fitness functions, i.e., we never compare $f^{(t_1)}(x)$ with $f^{(t_2)}(x')$ for different $t_1 \neq t_2$. In other words, the fitness of all individuals changes in each round. Since this requires $\mu+1$ function evaluations per generation, we define \emph{the runtime as the number of generations} until the algorithm finds the optimum. This deviates from the more standard convention to count the number of function evaluations (essentially by a factor $\mu+1$), but it makes the performance easier to compare with previous work on static linear functions. Also, note that the runtime equals the number of search points that are sampled, up to an additive $-(\mu-1)$ from initialization.

We consider two types of dynamic functions. A \emph{dynamic linear function} is described by a distribution $\mathcal D$ on $\R^+$. For the $t$-th round, we draw $n$ independent samples $W_1^{(t)},\ldots,W_n^{(t)}$ and set
\begin{equation}
f^{(t)}(x) := \textstyle\sum_{i=1}^{n} W_i^{(t)}\cdot x_i  .
\end{equation}
So $f^{(t)}$ is a positive with positive weights. Thus all $f^{(t)}$ share the same global optimum $1\ldots1$, have no other local optima, and they are monotone, i.e., flipping a zero-bit into a one-bit always increases the fitness.

For \emph{dynamic binval}, \dynbv, in the $t$-round we draw a permutation $\pi_t: \{1..n\} \rightarrow \{1..n\}$ uniformly at random, and define
\begin{equation}
f^{(t)}(x) = \textstyle\sum_{i=1}^{n} 2^i \cdot x_{\pi_t(i)}.    
\end{equation}
So, we randomly permute the bits of the string, and take the binary value of the permuted string. As for dynamic linear functions, all $f^{(t)}$ share the same global optimum $1\ldots1$, have no other local optima, and prefer one-bits to zero-bits.

\iftoggle{journal}
{We claim that in a certain sense, \dynbv is a limit case of dynamic linear functions in which the tail of the distribution $\mathcal D$ becomes infinitely heavy. Let us make this precise. Consider a noisy linear function $f$ and two strings $x$, $x'$ that differ in $k$ bits. To ease notation, assume they differ in the first $k \geq 2$ bits. The \emph {order statistics} $W_{(1)}^{(t)}\leq \ldots \leq W_{(k)}^{(t)}$ are obtained from $W_{1}^{(t)},\ldots,W_{k}^{(t)}$ by sorting, i.e., the first order statistics $W_{(1)}^{(t)}$ is the smallest value among $W_{1}^{(t)},\ldots,W_{k}^{(t)}$, and so on. If the distribution $\mathcal{D}$ is sufficiently skewed, then the probability 
\begin{equation}\label{eq:order-stat}
p_k := \mathbb{P}\Big(W_{(k)}^{(t)} > \sum\nolimits_{i=1}^{k-1} W_{(i)}^{(t)}\Big) 
\end{equation}
comes arbitrarily close to one. However, conditioned on the event in~\eqref{eq:order-stat}, comparison with respect to the dynamic linear function is \emph{equivalent} to comparison with respect to \dynbv. If we compute the difference $f^{(t)}(x)-f^{(t)}(x') = \sum_{i=1}^k W_i^{(t)}(x_i-x_i')$, then the sign of this difference will only depend on which of the $k$ bits has the highest weight, since this summand dominates the whole remaining sum. The position of the highest weight bit is uniformly at random, so effectively, conditioned on the event in~\eqref{eq:order-stat}, the dynamic linear function picks randomly one of the bits in which $x$ and $x'$ differ, and bases its comparison only on that bit. This is exactly what \dynbv also does.

In fact, this reasoning was implicitly used in~\cite[Theorem 7]{ulyjohan2018}. There, to construct a hard example for the \ooea with $c > c_0 \approx 1.59$, the authors used $\mathcal D$ as a Pareto distribution Par$(\beta)$, showed that for this distribution $p_k \geq (k-1)^{-\beta}$, and observed that this term comes arbitrarily close to $1$ as $\beta \to 0$. Then they picked $\beta$ so close to zero that the difference was negligible. Thus, in effect, they used for their hard function that \dynbv can be arbitrarily well approximated by dynamic linear functions, and their proof implies that the \ooea also needs exponential time for \dynbv if $c > c_0$, and time $O(n\log n)$ for $c <c_0$. Note that in general one needs to be a bit careful when two limits $n\to\infty$ and $\beta\to\infty$ are involved. However, this is no problem for the \ooea since there are strong tail bounds for the number of bits in which $x$ and $x'$ differ~\cite{ulyjohan2018}. The same argument also extends to populations in situations in which the population tends to degenerate to copies of a single points, as it is the case for $(\mu+1)$ algorithms if it is hard to find improvements~\cite{Dichotomy}.
}
{%if journal = false
In a certain sense, \dynbv is a limit case of dynamic linear functions, and it was implicitly used in~\cite[Theorem 7]{ulyjohan2018} to construct a hard instance of dynamic linear functions. Due to space constraints, we refer the reader to the full version~\cite{fullversion} for a discussion.
}

\subsection{Runtime Simulations}

Recall that we count the runtime as the number of generations until the optimum is sampled. We run the different algorithms to observe the distribution of the runtime. A run terminates if either the optimum is found, or an upper limit of generations is reached. Unless otherwise noted, the upper limit is set to be $100 e^c/c\cdot n  \ln{n}$, which is $100$ times larger than the expected runtime of the \ooea~\cite{witt2013tight}. The python code of our running time simulation can be found in our GitHub repository~\cite{Github}. Unless otherwise noted, each data point is obtained by $30$ independent runs. 
\iftoggle{journal}
{

To verify the correctness of our simulation we first measure mean and variance of the runtime of the (1+1)-EA with mutation parameter $c = 1$ on the function \onemax (the linear function where all weights are $1$), and compare them with the highly accurate values derived in~\cite{OneMaxRuntime}. We compute mean and variance of 3000 runs and find that our observed mean deviates $0.16\%$ of the predicted mean, while the observed variance of is within $2.3\%$ of the predicted variance.
}
{%if journal = false
We have verified correctness of our implementation by experiments with the \ooea, see the full version~\cite{fullversion} for details. 
}

To visualize runtimes, we use plots provided by the IOHprofiler~\cite{IOHprofiler}. 
\iftoggle{journal}
{
As an example, the runtime of the \ooea with mutation parameter $c = 1.0$ on \onemax is visualized in Figure~\ref{fig:OneMaxRuntime}. 
}
{% if journal = false
}
Note that time (i.e., number of generations) is displayed \emph{on the y-axis}, while the x-axis corresponds gives the number of 1-bits. Thus, a steep part of the curve corresponds to slow progress, while a flat part of a curve corresponds to fast progress. Also, mind that the y-axis uses log scale.

\iftoggle{journal}
{\begin{figure}[h]
    \centering
    \includegraphics[width=5cm]{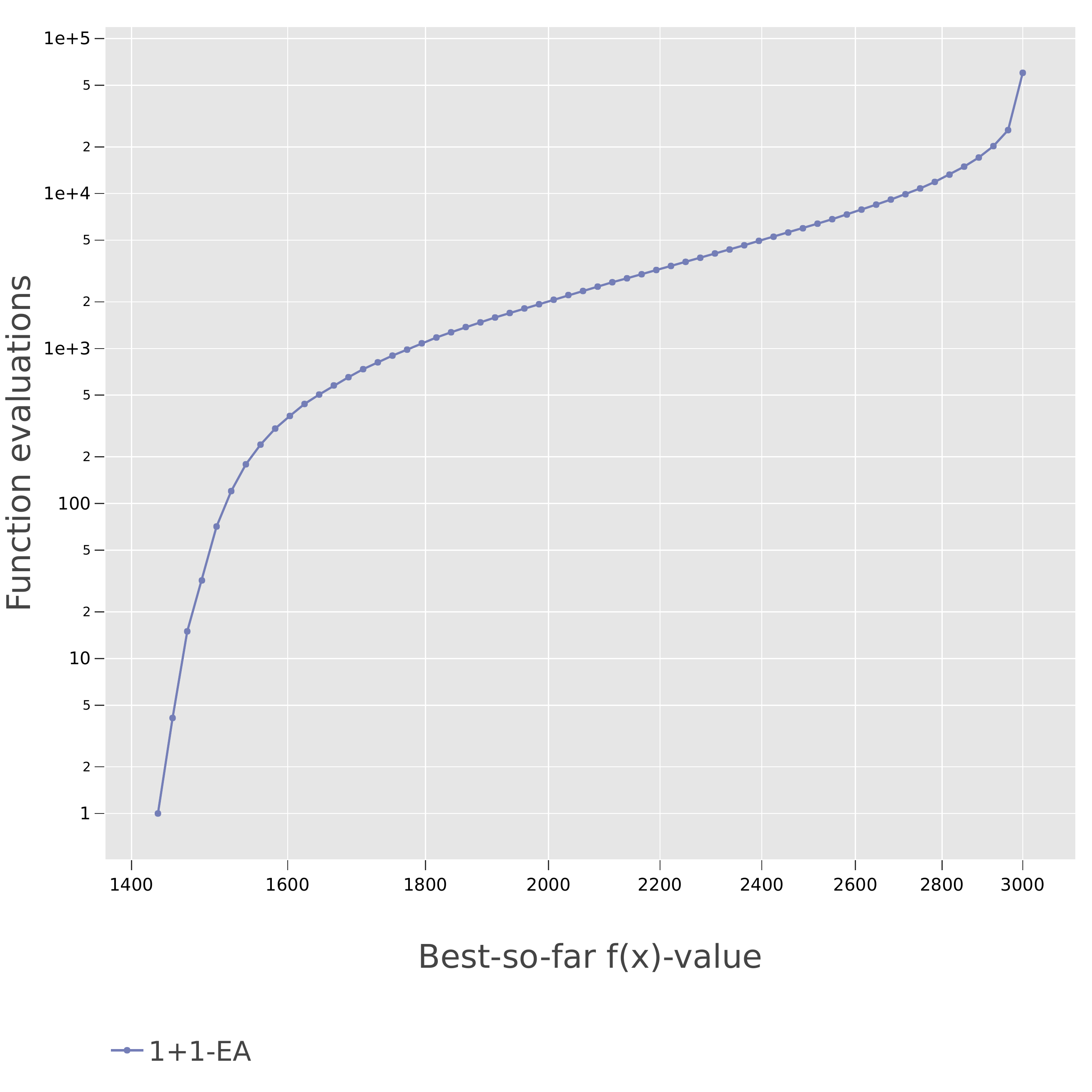}
    \caption{Runtime of the \ooea on \onemax}
    \label{fig:OneMaxRuntime}
\end{figure}
}
{% if journal = false
}

Due to the exponential runtimes, we frequently encounter the problem that runs are terminated due to the iteration limit of $100 e^c/c\cdot n\ln{n}$ generations. 
\iftoggle{journal}
{In this case we will often plot two values, which are a lower and upper estimates of for the expected runtime. The lower bound is the \emph{mean runtime}, i.e., the average number of generations among the \emph{successful} runs. By definition, this number never exceeds the upper limit. For the upper bound, we use the \emph{expected runtime} (ERT) as defined by the IOH profiler. The ERT is calculated by drawing random runs from our pool of runs, until a successful run is drawn. Then the runtime of all drawn runs is added up, and the expectation of this process is defined as the ERT. This estimates the expected runtime if we start over the algorithm every time the iteration limit is reached. The ERT overestimates the runtime if the state at hitting the iteration limit is better then the starting state of a restart, which is the case in all benchmarks we consider. (It may not be the case for deceptive functions.) We remark that if \emph{all} runs hit the iteration limit, then our lower bound (mean runtime) is the iteration limit, while the upper bound (ERT) is infinity. A more detailed explanation of the ERT can be found in~\cite{IOHprofiler}.
}
{% if journal = false
In this case, we will often plot two values: the \emph{mean runtime} is a lower bound estimate for the actual runtime, while the \emph{expected runtime} (ERT) as defined by the IOH profiler is an upper bound. For a definition and a discussion of both, see the full version~\cite{fullversion} and~\cite{IOHprofiler}.
}

%We will often show the expected running time (ERT) as well as the mean. If all our runs managed to find the optimum before they maximum number of iterations was reached, ERT and mean will agree. Then, the average time it took to find a string with a certain number of 1-bits is plotted. If, however, some runs were unsuccessful in finding the optimum within the maximum number of iterations, ERT and mean will differ. The mean will just be taken from the runs that were successful, which is underestimating the runtime as these are exactly the "good" runs. The ERT is calculated by drawing random runs from our pool of runs, until a successful run is drawn. Then the runtime of all drawn runs is added up to give the ERT. Typically, we will simulate an algorithm between 10 and 30 times, so the pool of runs would then consist of 10 to 30 runs, some of which might not have been successful in finding the optimum (unsuccessful runs reached the iteration limit before finding the optimum). This essentially amounts to the runtime as if we started over the algorithm every time the iteration limit is reached. The ERT will overestimate the runtime in our case, as restarting will almost never be beneficial. We will regularly plot both quantities, to give as complete a picture as possible. A more detailed explanation of mean and ERT can be found in \cite{IOHprofiler}.

\noindent \textbf{Comparison of Runtimes.} We want to compare runtimes for different algorithms and values of $c$. We denote by $R_{c}^{\text{Alg}}$ the random variable describing the runtime of Algorithm Alg for a specific $c$. Because our sample size is fairly small (10 to 30), we compare runtimes using the Wilcoxon-Mann-Whitney test. The Wilcoxon-Mann-Whitney test is a test of the null hypothesis that with probability at least $1/2$ a randomly selected value from one runtime distribution will be at most (at least) a randomly selected value from a second runtime distribution. A small p-value would then indicate that the runtime of one algorithm, treated as random variable, is larger (smaller) than the runtime of the other algorithm in significantly more than half of the cases. 

We will also be interested in quantifying \emph{by how much} an algorithm is slower than another algorithm. To this end, we will determine the largest factor $d \geq 1$  by which we can multiply one runtime distribution such that the Wilcoxon-Mann-Whitney test still yields a statistically significant p-value. For example, we will find that even if we multiply the runtime $R_{2.0}^{\toea}$ with $d=63.15$ then this is still significantly faster than $R_{2.5}^{\toea}$ according to the Wilcoxon-Mann-Whitney test. Note that this is a posteriori hypothesis since the factor $d$ is chosen in hindsight. Therefore, it must not be treated as an actually significant result. Still, it gives useful information, and shows that $R_{2.0}^{\toea}$ is \emph{very much} smaller than $R_{2.5}^{\toea}$. All tests are done with R.

% in section 4.1, we compare the (1+1)-EA for different values of $c$. Specifically, we compare $R_{2.0}^{(1+1)-EA}$ to $R_{1.5}^{(1+1)-EA}$, expecting $R_{2.0}^{(1+1)-EA}$ to be much greater than $R_{1.5}^{(1+1)-EA}$. To quantify how much greater, the running time for $c = 2.0$ is compared to $c = 1.5$, i.e we test $R_{2.0}^{(1+1)-EA}$ against $d \cdot R_{1.5}^{(1+1)-EA}$ and gradually improve $d$ until the Wilcoxon-Mann-Whitney-Test stops giving significant p-values. We then state the largest $d \geq 1$, s.t the p-value was still significant, i.e. $<0.05$. In this case , the largest such $d$ is $d = 38.84$, indicating that even if we multiply the runtime distribution for $c=1.5$ by $38.84$, the distribution for $c=2.0$ is still significantly larger. The tests are done with R.

\subsection{Analysis of Population Drift}

Recall that $X_i$ was defined to be the number of zero-bits in an individual of the i-th degenerated population, where degenerate means that all individuals are copies of the same search point. To be precise, if $P^{(i)}$ is the $i$-th degenerate population, then $P^{(i+1)}$ is the first degenerate population after $P^{(i)}$ has changed at least once. That does not exclude the possibility $P^{(i)} = P^{(i+1)}$, but we require at least one intermediate step in which an offspring is accepted that is not a copy of the parent(s) in $P^{(i)}$. Then the \emph{degenerate population drift} (population drift for short) is $\mathbf{E}[X_i - X_{i+1}\mid X_{i} = y]$. 
\iftoggle{journal}
{We will estimate this drift with Monte-Carlo simulations. Moreover, for $y=o(n)$ we will derive a Markov chain state diagram in which all transition probabilities coincide with the transition probabilities in the real process up to $(1+o(1))$ factors. By analyzing this state diagram, we are able to compute the population drift up to minor order error terms. 
}
{%if journal = false
We will estimate this drift with Monte-Carlo simulations. Our code is publicly available~\cite{Github}. Moreover, the full version~\cite{fullversion} contains an exact asymptotic formula for the population drift of the \toea and \toga in the limit $n\to \infty$ and $y =o(n)$, derived by Markov chain analysis. We omit them for space reasons. 
}
\iftoggle{journal}{

To motivate the use of population drift, consider the following example for $\mu=2$. Take a population $\{x_1,x_2\}$, and assume that $x_1$ has at least as many one-bits than $x_2$. Then in every iteration, there is a chance to simply copy $x_1$ (mutate but flip none of the bits), and accept it. For this to happen, we need to first mutate $x_1$, flip no bits at all and accept the new offspring. The probability of mutating $x_1$ and flipping no bits is given by $\frac{1}{2} \cdot (1 - \frac{c}{n})^n \approx e^{-c}/2$ for the \toea and $\frac{1}{4} \cdot (1 - \frac{c}{n})^n \approx e^{-c}/4$ for the \toga and the \togan. The probability of accepting $x_1$ is at least $1/2$, as $x_1$ has at least as many 1s as $x_2$. Hence, we have a constant probability to degenerate to a population $\{x_1,x_1\}$ in every iteration. This implies that any population degenerates within an expected constant number of rounds. The argument can be generalized to larger $\mu$, see~\cite{Dichotomy}: for every constant $\mu$ and $c$, the expected time until the population degenerates from any starting population is $O(1)$. 
}
{%if journal = false
It is easy to see~\cite{Dichotomy} that for every constant $\mu$ and $c$, the expected time for degeneration is $O(1)$. 
}
Moreover, it was shown in~\cite{Dichotomy} that if the population drift is negative for $y = \alpha n$ for some $\alpha \in (1/2,1)$ then asymptotically the runtime is exponential in $n$. On the other hand, if the population drift is positive for all $\alpha$ then the runtime is $O(n \log n)$. Hence, we are trying to identify parameter regimes for which areas of negative population drift occur.

\section{Results}

\subsection{Runtimes}

%Inspired by the threshold results that was proven in \cite{ulyjohan2018} for the (1+1)-EA and empirically verifying that it holds not only for noisy linear functions but also for \textit{dynamic binval}, we test if a similar situation can be found for the (2+1)-EA. After using our different tools to investigate the behavior of the (2+1)-EA, it seems that indeed we are looking at a situation reminiscent of the (1+1)-EA.

\begin{figure}[H]
    \centering
    \includegraphics[width = 12cm]{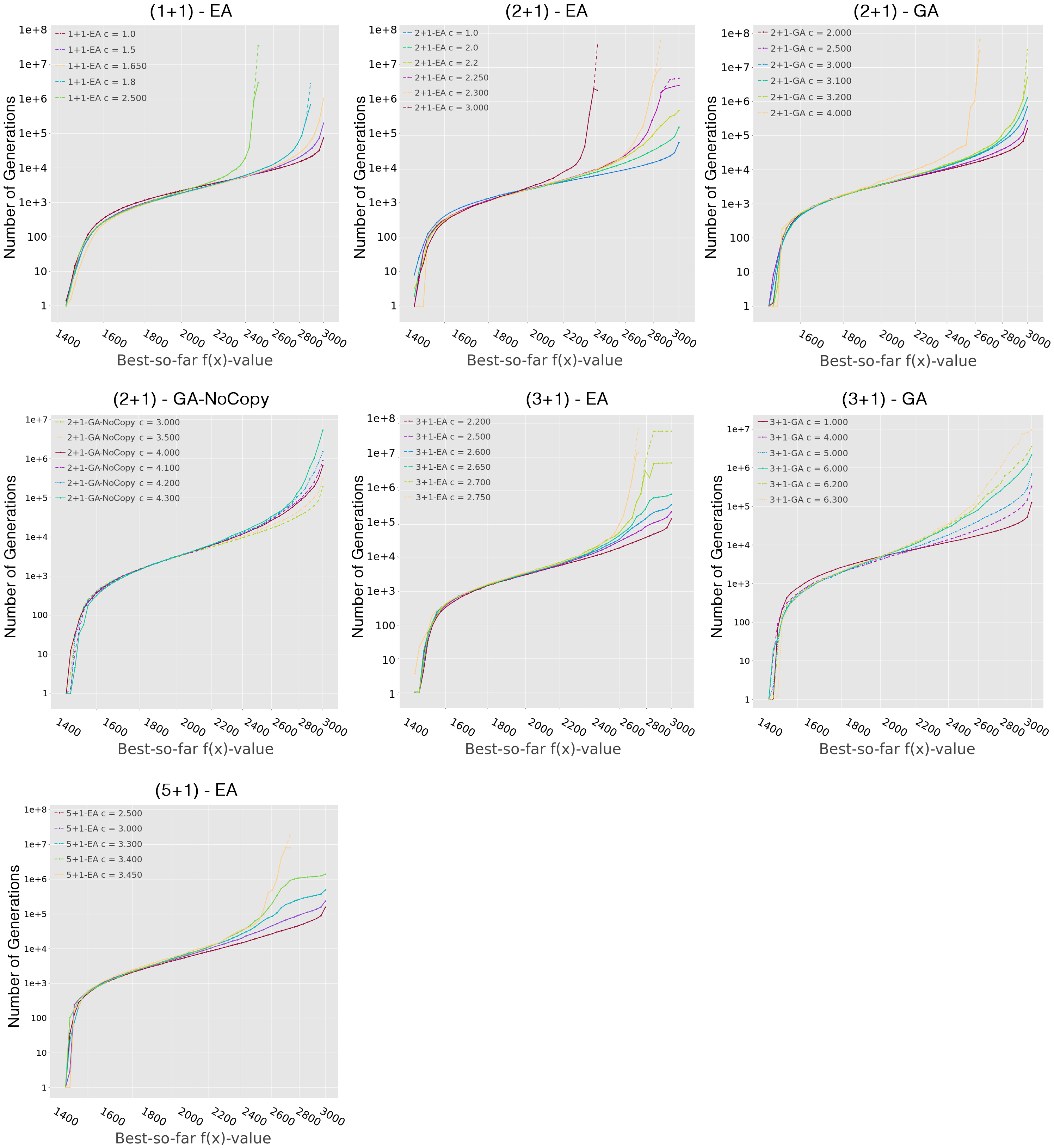}
    \caption{Comparison of runtimes for different algorithms and values of $c$. We choose values of $c$ that lie around the threshold for the respective algorithm. We plot both ERT and mean (both in the same color) if they differ significantly (mean is always the lower curve, see Section 3.1 for a more detailed explanation). }
    \label{fig:Runtimes}
\end{figure}

%\subsubsection{Runtime}
The results of our runtime simulations for different algorithms and values of $c$ can be found in Figures~\ref{fig:Runtimes}. As expected, we find a threshold behavior, i.e., there is a value $c_0$ such that the runtime increases dramatically as $c$ crosses this threshold. For the \ooea, by visual inspection we observe a threshold in $c_0 \in [1.5,1.8]$ (in agreement with the theoretically derived threshold $c_0 \approx 1.59$ from~\cite{ulyjohan2018}). For the \toea, it seems to be within $c_0 \in [2.2,2.3]$, for the \toga in $[3,3.2]$ and for the $\togan$ in $[4.1,4.3]$. Thus we obtain a clear ranking of the algorithms for $\mu \le 2$, which is \ooea (worst), \toea, \toga, and \togan. For the \thoea, the threshold appears to lie in the interval $[2.5,2.7]$, for the \thoga in $[6.0,6.3]$ and for the \fiveoea in $[3.3,3.45]$. We were not able to find a threshold behavior for the \fiveoga for any $c < 20$. These results further confirm that the GA variants are performing massively better than its EA counterparts. Moreover, both for EAs and GAs, a larger population size shifts the threshold to the right. For GAs, this is analogous to theoretical results for monotone functions, but for EAs the effect goes in the opposite direction than for monotone functions, see the discussion in Section~\ref{sec:intro}.

To validate the ranking for the algorithms with $\mu\le 2$ statistically, we use the following comparisons. If we compare $R_{2.0}^{\ooea}$ to $d \cdot R_{2.0}^{\toea}$ the Wilcoxon-Mann-Whitney test yields significant p-values $\leq 0.05$ for every $d \leq 57.88$. We conclude that for mutation parameter $c=2.0$, the \ooea is much slower than the \toea. In the same manner, $d \cdot R_{2.5}^{\toga}$ is  significantly smaller than $R_{2.5}^{\toea}$ for $d \leq 39.09$, and $d \cdot R_{3.5}^{\togan}$ is significantly smaller than $R_{3.5}^{\toga}$ for $d \leq 63.36$. This confirms the aforementioned ranking of the algorithms.

To establish intervals for the critical value $c_0$ of the algorithms with $\mu \leq 2$, we compare $R_{2.0}^{\ooea}$ to $d \cdot R_{1.5}^{\ooea}$, and find that the latter is significantly smaller for all $d \leq 38.84$. We interpret this huge drop in performance as strong indication that the threshold lies in the interval $c_0\in[1.5,2]$. Likewise, $R_{2.5}^{\toea}$ is larger than $d \cdot R_{2.0}^{\toea}$ for $d \leq 63.15$, $R_{3.5}^{\toga}$ is larger than $d \cdot R_{3.0}^{\toga}$ for $d\leq 29.00$, and $R_{4.5}^{\togan}$ is larger than $d \cdot R_{4.0}^{\togan}$ for $d \leq 29.59$, all with $p<0.05$.

%If we compare $R_{2.5}^{\toea}$ to $d \cdot R_{2.0}^{\toea}$, the Wilcoxon-Mann-Whitney test yields significant p-values for every $1 \leq d \leq 63.15$, indicating that the \toea is significantly faster for mutation parameter $c=2.0$ than for $c = 2.5$. Comparing $R_{2.0}^{\toea}$ to $d \cdot R_{2.0}^{\ooea}$ yields significant p-values for $d \leq 57.88$. We conclude that for mutation parameter $c=2.0$, the \ooea is much slower than the \toea. Therefore the threshold for the \toea seems to be substantially higher than for the \ooea.

\subsubsection{Degenerate population drift}
\begin{figure}[h]
    \centering
    \includegraphics[width=12cm]{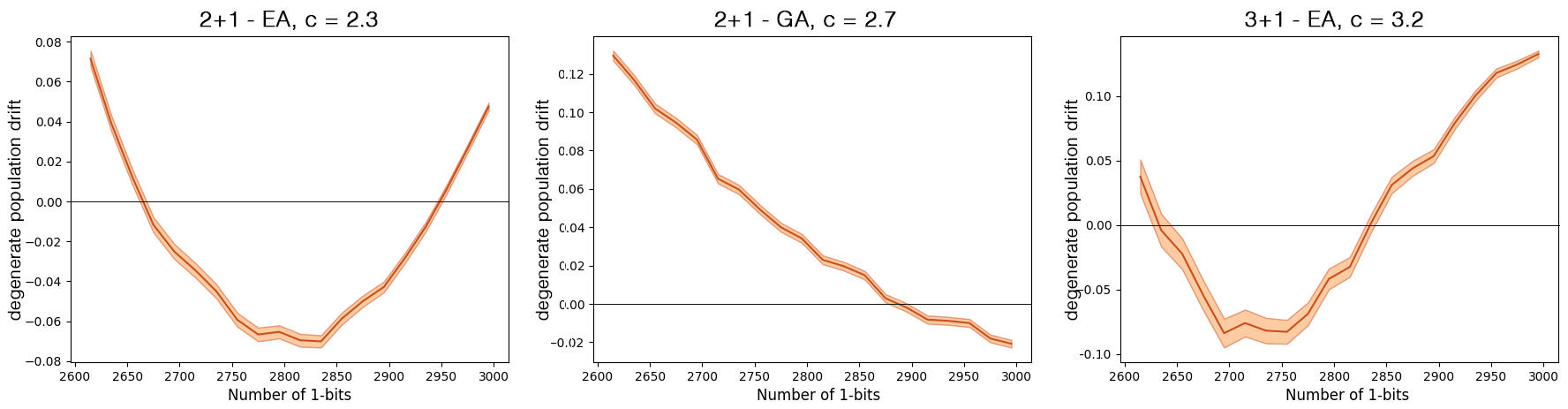}
    \caption{Degenerate population drift for different algorithms and values of $c$ just above the respective thresholds. The shaded area shows standard deviation.}
    \label{fig:Drift}
\end{figure}

We estimate the degenerate population drift by Monte-Carlo simulation on the \toea with $c = 2.3$, which is slightly above the threshold. The results are visualized in Figure \ref{fig:Drift}. We can clearly see that the conditional population drift is negative in the area between 300 and 50 one-bits away from the optimum, but then becomes positive again when being less than 50 one-bits away from the optimum. We conclude that the hardest part for the \toea is not around the optimum. We obtained similar results for the \thoea, also visualized in Figure \ref{fig:Drift}. This surprising result is similar to monotone functions~\cite{Dichotomy,ExpSlowdown}, see the discussion in Section~\ref{sec:intro}. 
%behavior was only recently observed for the first time in \cite{ExpSlowdown}, on monotone functions. There it was shown that for every $c$, we can make $\mu$ large enough s.t the ($\mu$+1)-EA needs exponential time and this slowdown is happening far away from the optimum. As an interesting novelty, while in \cite{ExpSlowdown} the effect was only proven for sufficiently large $\mu$, we are able to observe it already for $\mu = 2$.

For the \toga, the picture looks entirely different, as the drift is now strictly decreasing. We can only observe a negative drift area right at the optimum, starting from about 2900 1-bits. This behavior is similar to the \ooea and the \mogan, where the most difficult part is also close to the optimum (data not shown). Unfortunately, due to the large value of $c_0$, we were not able to obtain a conclusive result for the \thoga within reasonable computation time. 

\begin{figure}
    \centering
    \includegraphics[width=12cm]{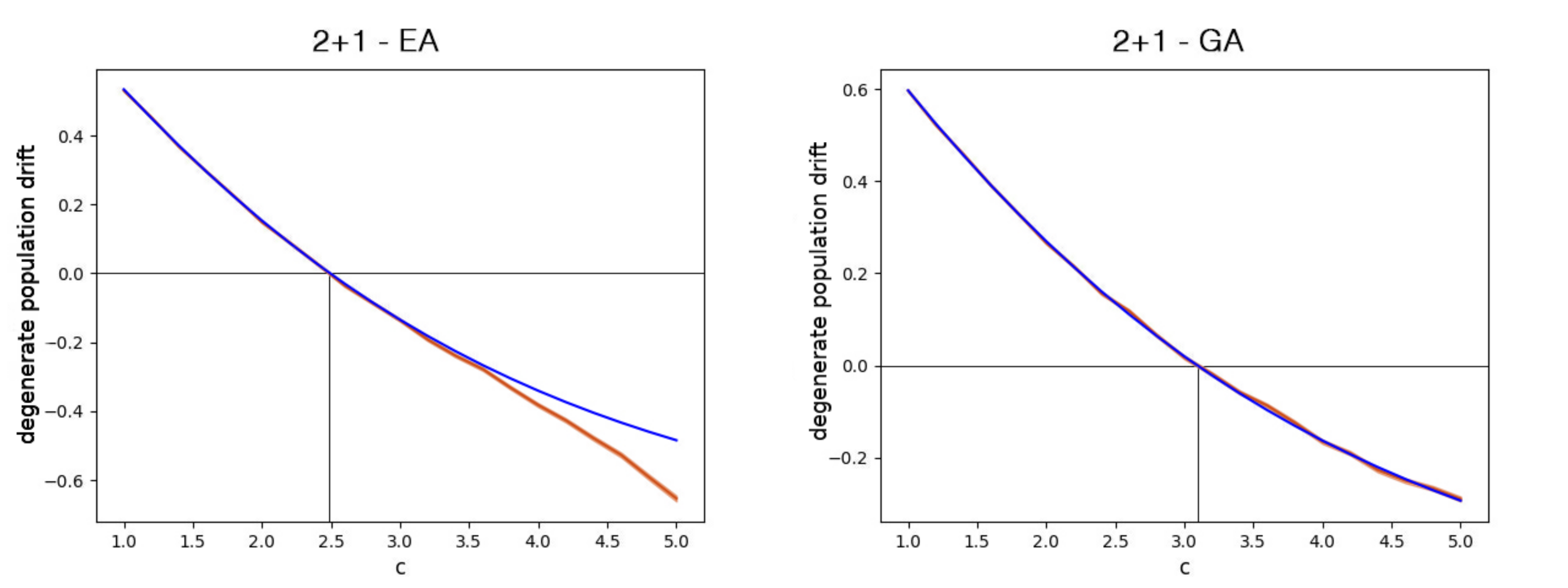}
    \caption{Degenerate Population Drift for the \toea and \toga at the optimum. Blue: asymptotic formula. Orange: Monte Carlo simulation.}
    \label{fig:State}
\end{figure}

For the \toea and \toga, we also derive exact asymptotic formulas for the population drift close to the optimum. 
\iftoggle{journal}
{We postpone the derivation to the appendix. 
}
{%if journal = false
For space reasons, we refer the reader to the full version~\cite{fullversion}. 
}
We compare the formula with the estimates via Monte Carlo simulation for different values of $c$, using $n = 3000, y = 1$, see Figure~\ref{fig:State}. We can see that the curves match closely for small $c$, and that we get a moderate fit for larger $c$. We suspect that the deviations for the \toea come from the expectation of the population drift being influenced by large but rare values of $X_i - X_{i+1}$ for large $c$. So the Monte Carlo simulations might be missing parts of this heavy tail. This tail is less heavy for the \toga, since the probability to produce duplicates is always high, and thus degeneration happens quickly even for large $c$. In both cases, the curves agree perfectly in the \emph{sign} of the population drift, which is our main interest. Negative drift at the optimum occurs at $c > 3.1$ for the \toga, which matches matches well the threshold obtained from runtime simulations. However, as expected, there is \emph{no such match} for the \toea, where the threshold for the population drift at the optimum is $2.5$ while the threshold for the runtime is below $2.3$. I.e., the \toea already struggles for values of $c$ for which optimization around the optimum is easy. This confirms that the hardest region for the \toea (but not the \toga) is not at the optimum, but a bit away from it. % (remember that using our simulations, we could observe very large running times and negative drift areas already for $c = 2.3$).

%Even though the tree is a bit more complicated than in the case of the \toea, the resulting expression that can be evaluated. We plot the population drift we compute in this way for $1 < c < 5$ and $y = 1$, and compare it to the drift obtained via Monte Carlo simulations (in orange), see Figure \ref{fig:2+1GADriftState}. We can see that the curves match closely for all depicted values of $c$. We observe that the drift becomes negative at about $c = 3.1$, which is the same value obtained from our runtime simulations. Note that this match occurs for the \toga, but not for the \toea. This confirms that for the \toga, but not for the \toea, the most difficult part lies at the optimum.

\section{Conclusions}
\label{sec:conclusions}

We have studied the effect of population size and crossover for the dynamic \dynbv benchmark. We have found that the algorithms generally profited from larger population size. Moreover, they profited strongly from crossover, even more so if we forbid crossovers between identical parents.

We have studied the case $\mu \le 2$ in more depth. Remarkably, there is a strong qualitative difference between the \toea and the \toga. While for the latter one, the hardest region for optimization is close to the optimum (as one would expect), the same is not true for the \toea. We believe that this is a interesting discovery. The only hint at such an effect on \onemax-like functions that we are aware of is for monotone functions~\cite{ExpSlowdown}. However, the results in~\cite{ExpSlowdown} predict that large population sizes hurt the \moea, in opposition to our findings. Currently we are lacking any understanding of whether this comes from the small values of $\mu$ that we considered here, or whether it is due to the differences between monotone and dynamic linear functions.

For future work, there are many natural questions. We have chosen the \moga to decide randomly between a mutation and a crossover step, but other choices are possible. Even with our formulation, it might be that the probability $1/2$ for choosing crossover has a strong impact. Also, we have exclusively focused on the limiting case \dynbv, but dynamic linear functions are also interesting for less extreme case weight distributions. Finally, an interesting variant of dynamic linear functions or \dynbv might not change the objective every round, but only every $s$ rounds (our runtime simulation already supports this feature and is publicly available). 

%\subsection*{Acknowledgments}
%Part of the work was inspired by discussions at the Dagstuhl meeting 19171 on Theory of Randomized Optimization Heuristics. In particular, we thank Benjamin Doerr for proposing to consider the \olea on monotone functions, which started this line of research. The study of monotone functions was fostered by the COST Action CA15140 ``Improving Applicability of Nature-Inspired Optimisation by Joining Theory and Practice'', and the author has proposed to include monotone functions into the set of benchmarks developed by working group 3. 

%\begin{thebibliography}{10}
%\providecommand{\url}[1]{\texttt{#1}}
%\providecommand{\urlprefix}{URL }
%\providecommand{\doi}[1]{https://doi.org/#1}
%
%\bibitem{doerr2016optimal}
%Doerr, B.: Optimal parameter settings for the (1+ $\lambda$, $\lambda$) genetic
%  algorithm. In: GECCO (2016)
%
%\end{thebibliography}

\iftoggle{journal}
{
\appendix

\section{Appendix}

\subsection{\toea State diagram of \toea}\label{sec:toeaMarkov}

Here, we derive an expression for the conditional drift $\E[X_i - X_{i+1}\mid X_{i} = y \in o(n) ]$ based on a state diagram for the \toea. Recall that we defined $X_i$ to be the number of zero-bits in the i-th degenerated population. 
We define $\mathcal{E}_{\text{progress}}$ as the event that an offspring is accepted into the degenerate population that is not identical with the old search point in the population. 
We will always assume that we are working close to the optimum, which means the number of 0-bits, $y$, is $o(n)$. This allows us to ignore cases where more than one 0-bit was flipped when going from one degenerated population to the next. Additionally, we are interested only in the case where $n \rightarrow \infty$. We will encounter several $o(1)$ terms in our calculations that will go to 0 as n becomes large.\\

We claim that the development of the population follows the state diagram in Figure~\ref{StateEADrift}. The transition probabilities are written below the arrows. Before we justify Figure~\ref{StateEADrift} in detail, note crucially that an arrow may summarize several generations. For example, assume that the algorithm generates a population $\{x,y\}$ in which $x$ strictly dominates $y$, i..e, every one-bit in $y$ is also a one-bit in $x$ (and the converse is not true). In particular, $x$ is strictly fitter than $y$. We could model this situation by a state $\tilde S$. However, recall that it takes only $O(1)$ steps until a copy of $x$ is generated. Since by our assumption $y=o(n)$, the probability that any zero-bit is flipped into a one-bit in this time is $o(1)$. Hence, whp (with high probability, i.e., with probability $1-o(1)$) all offspring are strictly dominated by $x$ until a copy of $x$ is created, and then the population degenerates. Hence, we would obtain an arrow from $\tilde S$ to the state $\{x,x\}$ with probability $1-o(1)$, and the $o(1)$ will only affect the minor order terms of the final result. In cases like this, we will not draw the state $\tilde S$ in the first place. Instead, we may omit it, and replace any arrow to $\tilde S$ with an arrow (of the same probability) to the state $\{x,x\}$. This will allow us to keep the state diagram simple.

The top vertex represents a degenerate starting state $\{x,x\}$. Let $y$ be the number of 0-bits in an individual $x$ of our initial degenerated population $\{x,x\}$. A superscript denotes the number of additional 1-bits of an individual compared to $x$. This number can also be negative. By $S(k)$ a state where the next degenerated population has $(y-k)$ zero-bits. So the initial top state is identical with $S(0)$.\\ 

%By definition of the degenerate population drift, we may condition on $\mathcal{E}_{\text{progress}}$ in the first step. 
Starting with a population $\{x,x\}$, three things can happen. First, no bits at all or only 1-bits are flipped. Then, we will surely reject the offspring $\overline{x}$, as it is dominated (there is no position where $\overline{x}$ has a 1-bit and $x$ a 0-bit) by $x$. When being close to the optimum, this is what happens most of the time, as we have only a couple of zero-bits left and they will rarely be flipped. \\
Secondly, we could also just flip a 0-bit, but no 1-bits. Then we are in a state such that the offspring $\overline{x}^1$ is dominating $x^0$. Assuming that no further 1-bits are flipped, the population will whp degenerate to $\{\overline{x},\overline{x}\}$, as no offspring will be accepted over $\overline{x}$. At some point, $\overline{x}$ will just be copied and accepted.\\
The third and perhaps most interesting case occurs when a 0-bit and $r \geq 1$ 1-bits are flipped. We then have an offspring $\overline{x}^{(1-r)}$ such that $\overline{x}$ and $x$ differ at exactly $r+1$ positions. In these $r+1$ positions, $\overline{x}$ has exactly one 1-bit, and 0-bits everywhere else. Now $\overline{x}$ will either be rejected, which results in the same population $\{x,x\}$ we started with, or will be accepted. If it is accepted, we land in a state which we called $F(r)$ (green in Figure \ref{StateEADrift}).\\
Starting from $F(r)$ we can either mutate $x$ or $\overline{x}$. Assume we mutate x, and flip $s$ 1-bits to create $\dot{x}^{-s}$. Notice that $\dot{x}$ is dominated by $x$. We can then either accept $\dot{x}$ to land in $S(0)$ or reject it to return to $F(r)$ once again.\\
If we mutate $\overline{x}$, we create an offspring $\overline{\overline{x}}^{1-r-s}$, which is dominated by $\overline{x}$. If we accept $\overline{\overline{x}}$, we will conclude in $S(1-r)$, otherwise we will go back to $F(r)$.

Putting it all together, we can get an explicit formula for the drift. In the diagram, note that we need to compute the drift conditioned on $\mathcal{E}_{\text{progress}}$, i.e., assuming we visit either $S(1)$ or $F(r)$. 

First, we compute the expected number of 0-bits when starting in state $F(r)$. Let us slightly abuse notation and also call this $F(r) := \E[X_{i+1} - X_i\mid X_i = y \in o(n)\ \land$ we are in state $F(r)$]. Recall that we land in state $F(r)$ if we flip one 1-bit and $r$ 1-bits to create an offspring $\overline{x}$ and accept it. Simply writing out the transition probabilities yields

\begin{equation*}
\begin{split}
F(r) = (1\pm o(1))\frac{1}{2}\sum_{s=0}^{n-y}& ((1-o(1)) \binom{n}{s} \Big(\frac{c}{n}\Big)^s \Big(1-\frac{c}{n}\Big)^{n-s} \cdot \frac{1}{r+s+1}\\
& \cdot \left((s+1) \cdot F(r) +  r \cdot 0 + (r+s)\cdot F(r) +  1 \cdot (1-r)  \right). 
\end{split}
\end{equation*}

Approximate the sums by letting them run up to infinity only give another factor $(1\pm o(1))$. The dominant terms have small $s$, and so we may approximate $\binom{n}{s}(c/n)^s(1-c/n)^{n-s} = (1\pm o(1))c^se^{-c}/s!$ in this case. Hence, we obtain
%Solving for $F(r)$ yields
%\begin{equation*}
%\begin{split}
%2F(r) = (&\sum_{s=0}^{\infty} \binom{n}{s} \cdot (\frac{c}{n})^s \cdot (1-\frac{c}{n})^{n-s} \cdot \frac{s+1}{r+s+1} \cdot F(r)+ \\
%    &\sum_{s=0}^{\infty} \binom{n}{s} \cdot (\frac{c}{n})^s \cdot (1-\frac{c}{n})^{n-s} \cdot \frac{r+s}{r+s+1} \cdot F(r) + \\
%    &\sum_{s=0}^{\infty} \binom{n}{s} \cdot (\frac{c}{n})^s \cdot (1-\frac{c}{n})^{n-s} \cdot \frac{1}{r+s+1} \cdot (1-r)) \cdot (1 + o(1)) \\
%  = (&\sum_{s=0}^{\infty} \binom{n}{s} \cdot (\frac{c}{n})^s \cdot (1-\frac{c}{n})^{n-s} \cdot (1 + \frac{s}{r+s+1}) \cdot F(r) + \\
%    &\sum_{s=0}^{\infty} \binom{n}{s} \cdot (\frac{c}{n})^s \cdot (1-\frac{c}{n})^{n-s} \cdot \frac{1}{r+s+1} \cdot (1-r)) \cdot (1 + o(1))  \\
%  = (&F(r) + \sum_{s=0}^{\infty} \binom{n}{s} \cdot (\frac{c}{n})^s \cdot (1-\frac{c}{n})^{n-s} \cdot \frac{s}{r+s+1} \cdot F(r) + \\
%    &\sum_{s=0}^{\infty} \binom{n}{s} \cdot (\frac{c}{n})^s \cdot (1-\frac{c}{n})^{n-s} \cdot \frac{1}{r+s+1} \cdot (1-r)) \cdot (1 + o(1)). \\
%\end{split}
%\end{equation*}
%
%We further approximate by replacing $(1-\frac{c}{n})^{n-s}$ by $(1 + o(1)) \cdot e^{-c}$ as the relevant terms have a small $s$. Additionally, we let $n \rightarrow \infty$, which allows us to get rid of the $o(1)$ terms. Also we expand the binomials and see that $\frac{n!}{(n-s)!} \cdot n^{-s} \rightarrow 1$ as $n \rightarrow \infty$
%
\begin{equation*}
\begin{split}
2F(r) = & (1\pm o(1))\sum_{s=0}^{\infty} \frac{c^s}{s!} \cdot e^{-c} \cdot \frac{r+2s+1}{r+s+1} \cdot F(r) + \sum_{s=0}^{\infty} \frac{c^s}{s!} \cdot e^{-c} \cdot \frac{1-r}{r+s+1}.\\
\end{split}
\end{equation*}

For the left hand side, we artificially write $F(r) = \sum_{i=0}^\infty c^se^{-c}/s! \cdot F(r)$. Solving for $F(r)$ yields
\begin{equation*}
\begin{split}
F(r) = (1\pm o(1))\frac{\sum_{s=0}^{\infty} \frac{c^s}{s!} \cdot e^{-c} \cdot \frac{1}{r+s+1} \cdot (1-r)}{\sum_{s=0}^{\infty} \frac{c^s}{s!} \cdot e^{-c} \cdot \frac{1}{r+s+1}\cdot (r+1)}= (1\pm o(1))\frac{1-r}{r+1}.
\end{split}
\end{equation*}

In order to compute the population drift, we need some elementary probabilities. Let $\mathcal{E}_0^r$ be the event that exactly $r$ zero-bits are flipped and $\mathcal{E}_1^r$ the event that exactly $r$ one-bits are flipped. Also, define $\mathcal{E}_{acc}$ to be the event that the offspring is accepted. Then
\begin{equation*}
\begin{split}
&\E[X_i - X_{i+1}\mid X_{i} = y ] = \frac{\mathbb{P}[\mathcal{E}_0^1 \land \mathcal{E}_1^0] + \sum_{r=1}^{r_{\max}}\mathbb{P}[\mathcal{E}_0^1 \land \mathcal{E}_1^r \land \mathcal{E}_{acc}] \cdot F(r)}{\mathbb{P}[\mathcal{E}_{\text{progress}}]},
\end{split}
\end{equation*}
where 
\begin{equation*}
\begin{split}
&\mathbb{P}[\mathcal{E}_0^0 \land \mathcal{E}_1^0] = (1-\frac{c}{n})^y. \\\\
&\mathbb{P}[\mathcal{E}_0^1 \land \mathcal{E}_1^0] = y \cdot \frac{c}{n} \cdot (1 - \frac{c}{n})^{n-1}.\\\\
&\mathbb{P}[\mathcal{E}_0^1 \land \mathcal{E}_1^r] = y \cdot \binom{n-y}{r} \cdot \frac{c}{n}^{r+1} \cdot (1-\frac{c}{n})^{n-r-1}.\\\\
&\mathbb{P}[\mathcal{E}_0^1 \land \neg \mathcal{E}_1^0 \land \neg \mathcal{E}_{acc}] = \sum_{r=1}^{r_{\max}} \mathbb{P}[\mathcal{E}_0^1 \land \mathcal{E}_1^r] \cdot \frac{r}{r+1}.\\\\
&\mathbb{P}[\mathcal{E}_0^1 \land \mathcal{E}_1^r \land \mathcal{E}_{acc}] = \mathbb{P}[\mathcal{E}_0^1 \land \mathcal{E}_1^r] \cdot \frac{1}{r+1}.\\\\
&\mathbb{P}[\mathcal{E}_{\text{progress}}] = 1 - \mathbb{P}[\mathcal{E}_0^0 \land \mathcal{E}_1^0] + \mathbb{P}[\mathcal{E}_0^1 \land \neg \mathcal{E}_1^0 \land \neg \mathcal{E}_{acc}].
\end{split}
\end{equation*}
Here, $r_{\max}$ could be as large as $n-y$, but for evaluation of the formula we will cut off at a value such that the difference is negligible, e.g. 50. %Now we are able to write down a formula for the  drift.

\begin{figure}[H]
\centering
\label{EADrift}
\begin{tikzpicture} [sloped,edge from parent/.style={draw,-latex}]
\node[bag, draw, circle, left = 3cm] {$x^{+0}$ \\ $x^{+0}$}
    child{
        node[bag, draw, circle,left = 0cm] {$x^{0}$ \\ $x^{0}$ \\ $\overline{x}^{(1-r)}$}
        child{
        node[bag, draw, circle,fill = yellow, right = 1cm] {$\mathbf{S(0)}$\\ $\neg \mathcal{E}_{\text{progress}}$}
    edge from parent
        node[above] {$f(x) > f(\overline{x})$}
        node[below] {$\frac{r}{r+1}$}
        }
        child{
        node[bag, draw, circle, right = 4cm, fill = green] (Node1) {$\mathbf{F(r)}$ \\ $x^{+0}$ $\overline{x}^{(1-r)}$}
            child{
            node[bag, draw, circle,above left= 0cm and 2cm] (Node2) {$x^{+0}$ \\ $\overline{x}^{1-r}$ \\ $\dot{x}^{-s}$}
                child{
                node[bag, draw, circle, fill = yellow, above = 2cm] {$\mathbf{S(0)}$}
            edge from parent
            node[above] {$\dot{x} > \overline{x}$}
            node[below] {$\frac{r}{r+s+1}$}
                }
        edge from parent
            node[above] {mutate $x$, flip s 1-bits}
            node[below] {$\frac{1}{2}\cdot (1-o(1))\cdot \binom{n}{s}\cdot (\frac{c}{n})^s\cdot (1-\frac{c}{n})^{n-s}$}
            }
            child{
            node[bag, draw, circle, above right= 0cm and 2cm] (Node3) {$x^{0}$ \\ $\overline{x}^{1-r}$ \\ $\overline{\overline{x}}^{1-r-s}$}
                child{
                node[bag, draw, circle, fill = yellow, above = 2cm] {$\mathbf{S(1-r)}$}
            edge from parent
            node[above] {$\overline{\overline{x}} > x$}
            node[below] {$\frac{1}{r+s+1}$}
                }
        edge from parent
            node[above] {mutate $\overline{x}$, flip s 1-bits}
            node[below] {$\frac{1}{2}\cdot (1-o(1))\cdot \binom{n}{s}\cdot (\frac{c}{n})^s\cdot (1-\frac{c}{n})^{n-s}$}
            }
    edge from parent
        node[above] {$f(x) < f(\overline{x})$}
        node[below] {$\frac{1}{r+1}$}
        }
    edge from parent
        node[above] {\begin{tabular}{c} 1 0-bit and $r$ 1-bits flipped    \end{tabular}}
        node[below]  {$y \cdot \binom{n-y}{r} \cdot(\frac{c}{n})^{r+1} \cdot (1-\frac{c}{n})^{n-r-1}$}
    }
    child{
        node[bag, draw, circle, below = 1cm, fill = yellow] (Test2) {$\mathbf{S(1)}$}
    edge from parent
        node[above] {\begin{tabular}{c} Exactly 1 0-bit and \\ no 1-bits flipped \end{tabular}}
        node[below]  {$y \cdot \frac{c}{n} \cdot (1-\frac{c}{n})^{n-1}$}
    }
    child{
        node[bag, draw, circle, right = -3cm, above = 0cm, fill = yellow] {$\mathbf{S(0)}$\\ $\neg \mathcal{E}_{\text{progress}}$}
    edge from parent
        node[above] {\begin{tabular}{c} No bits at all \\ or only 1-bits flipped \end{tabular}}
        node[below]  {$(1-\frac{c}{n})^y$}
    };
    
    \draw (Node2) [out = 90, in = 180, -latex]  to (Node1) node[below left = 1cm and 3.5cm] {$\overline{x} > \dot{x}$} node[below left=1cm and 2cm] {$\frac{s+1}{r+s+1}$};
    \draw (Node3) [out = 90, in = 0, -latex] to (Node1) node[below right= 1cm and 3.5cm] {$x > \overline{\overline{x}}$} node[below right =1cm and 2cm] {$\frac{r+s}{r+s+1}$};
\end{tikzpicture}
\caption{Transition diagram of a degenerated population in the \toea}
\label{StateEADrift}
\end{figure}
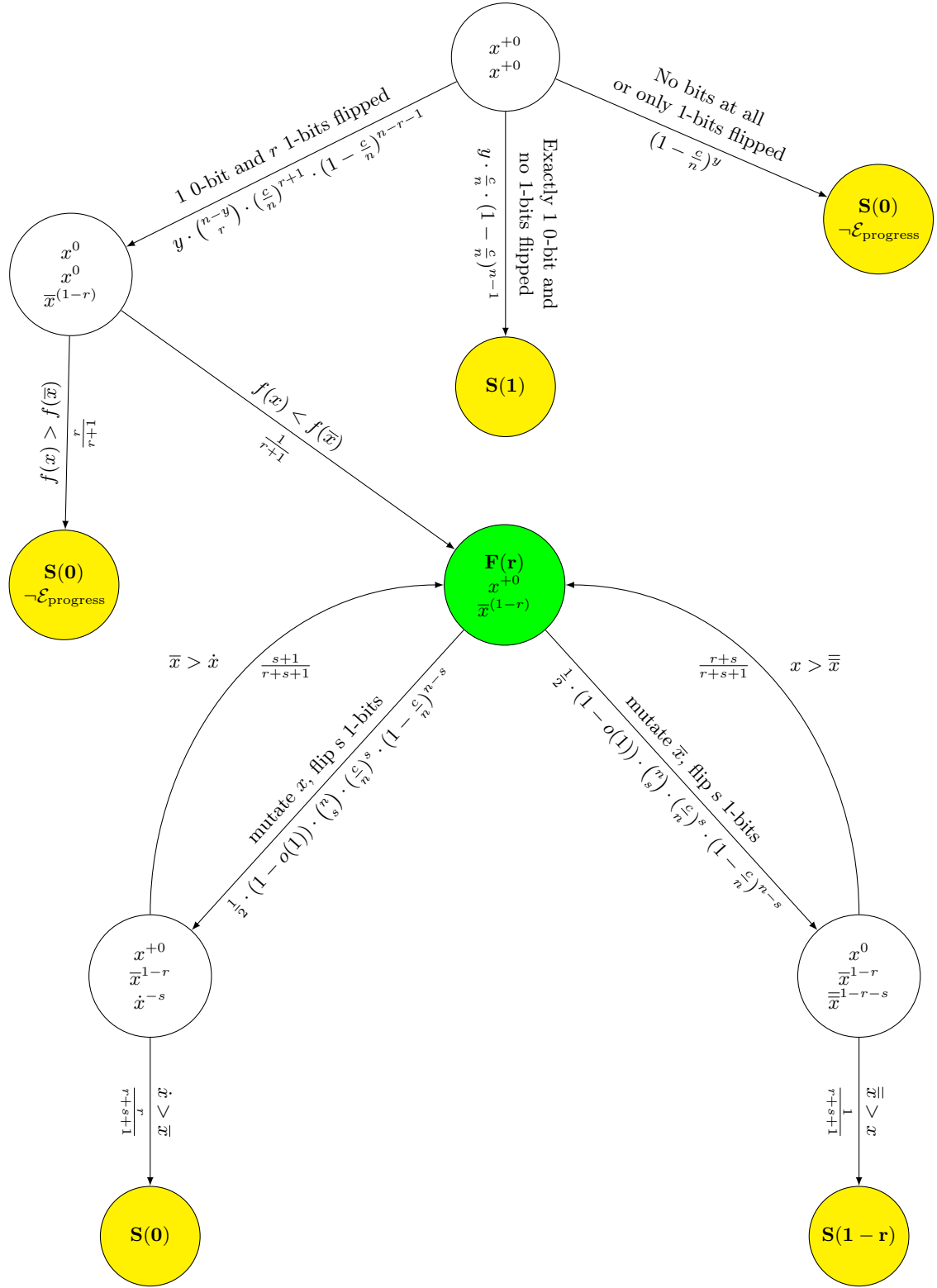

\section{\toga Population Drift}\label{sec:togaMarkov}

We compute the degenerate population drift at the optimum using a state diagram as we did with the \toea, see Figure~\ref{StateGADrift}. We only show the part that has changed significantly, which is the state that we now call $\overline{F}(r)$ (before $F(r)$). The upper part of the state diagram would be the same as in Figure \ref{StateEADrift}, except that now we can also do a crossover in the initial state, which will not change our population. Notice that the population drift will also exclude this case, since we only consider cases where at least one new offspring is accepted in the process.

Starting from $\overline{F}(r)$ we can do a mutation, which is exactly the same as in the \toea. Otherwise, we do a crossover between two strings. If we do a crossover between $x$ and $x$ ($x$ will just be copied in this case), we can either accept $x$ over $\overline{x}$, to get to a state $S(0)$, or reject it to return to $\overline{F}(r)$. Similarly, we can do a crossover between $\overline{x}$ and $\overline{x}$ and either accept to end in $S(1-r)$ or reject to land back in $\overline{F}(r)$.

Finally, we can do a crossover between $x$ and $\overline{x}$. To simplify notation, let us remove the bits on which $x$ and $\overline{x}$ agree. Moreover, we may assume that the first position is the one at which $\overline{x}$ has a one-bit, bit not $x$. So we are left with $x' = (0,1,1,\dots,1)$ and $\overline{x}' = (1,0,0,\dots,0)$. We can now do a case distinction depending on the shape of our crossover result $\dot{x}$. If $\dot{x}$ gets every one-bit, we land in a state $S(1)$, as $\dot{x}$ dominates both $x$ and $\overline{x}$. If $\dot{x}$ has a one-bit at the first position, plus inherits $s < r$ 1-bits from $x$, we can either remove $\overline{x}$ to go into a state $\overline{F}(r-s)$ or remove $x$, and land in $S(s+1-r)$, as $\dot{x}$ dominates $\overline{x}$. Finally, $\dot{x}$ could have a zero-bit at the first position and inherit $s$ one-bits from $x$. Then we could reject $\dot{x}$ to return to $\overline{F}(r)$ or accept it over $\overline{x}$ to conclude in $S(0)$.

We can also derive the conditional drift for the (2+1)-GA in exactly the same fashion as before. We start again by computing $\overline{F}(r) := \E[X_{i+1} \mid X_i = y \in o(n) \land \text{we are in state } \overline{F}(r)]$. Notice that now we obtain a recursive formula as we can go from a state $F(r)$ to a state $F(r-s)$.

\begin{equation*}
\begin{split}
\overline{F}(r) &= \frac{1}{4}\sum_{s=0}^{n-y} ((1\pm o(1)) \binom{n}{s}  \Big(\frac{c}{n}\Big)^s \Big(1-\frac{c}{n}\Big)^{n-s}\frac{1}{r+s+1} \\
    & \qquad \qquad\cdot \left((s+1) \cdot \overline F(r) + r \cdot 0 + (r+s)\cdot \overline F(r) + 1 \cdot (1-r)\right)\\
    & + \frac{1}{8} \cdot \frac{1}{r+1}\cdot \left(r \cdot 0 + 1\cdot \overline{F}(r) + r \cdot (1-r) + r \cdot \overline{F}(r) \right) \\
    &+ \frac{1}{4} \cdot \sum_{s=0}^{r-1} \binom{r}{s} \left(\frac{1}{2}\right)^{r+1} \cdot \frac{1}{r+1} \cdot \left(1\cdot (s+1-r) + r \cdot \overline{F}(r-s)\right)\\
    &+ \frac{1}{4} \cdot \left(\frac{1}{2}\right)^{r+1} + \frac{1}{4} \cdot \sum_{s=0}^r \binom{r}{s} \left(\frac{1}{2}\right)^{r+1} \frac{1}{s+1} \cdot \left(s\cdot 0 + 1 \cdot \overline{F}(r)\right).
\end{split}
\end{equation*}

As for the \toea, we may approximate $\binom{n}{s}(c/n)^s(1-c/n)^{n-s} = (1\pm o(1))c^se^{-c}/s!$ for large $s$, which are dominating. After simplification, we obtain
\begin{equation*}
\begin{split}
4\overline{F}(r) &= \sum_{s=0}^{\infty} \frac{c^s}{s!}  e^{-c} \cdot \left(\frac{r+2s+1}{r+s+1} \cdot \overline{F}(r) + \frac{1-r}{r+s+1}\right)\\
&+\frac{1}{2} \cdot \overline{F}(r) +\frac{1}{2} \cdot \frac{1-r}{r+1} +\sum_{s=0}^{r-1}\binom{r}{s}\left(\frac{1}{2}\right)^{r+1}\cdot\frac{s+1-r}{r+1}\\
&+\sum_{s=1}^{r-1}\binom{r}{s}\left(\frac{1}{2}\right)^{r+1}\cdot\frac{r}{r+1}\cdot \overline{F}(r-s)+\left(\frac{1}{2}\right)^{r+1}\cdot\frac{r}{r+1}\cdot\overline{F}(r)+\\
&+\left(\frac{1}{2}\right)^{r+1}+\sum_{s=0}^r \binom{r}{s} \cdot \left(\frac{1}{2}\right)^{r+1} \cdot \frac{\overline{F}(r)}{s+1}. 
\end{split}
\end{equation*}
Unfortunately, the formula does not simplify as much. Solving for $\overline{F}(r)$ yields a recursive formula:
\begin{equation*}
\hspace*{-1cm}
\begin{split}
&\overline{F}(r) = \\
&\frac{\sum\limits_{s=0}^{\infty}\frac{c^s}{s!} e^{-c}\frac{1-r}{r+s+1} + \frac{1-r}{2r+2} + \sum\limits_{s=1}^{r-1}\binom{r}{s}\left(\frac{1}{2}\right)^{r+1}\frac{r  \overline{F}(r-s)}{r+1} + \left(\frac{1}{2}\right)^{r+1} +  \sum\limits_{s=0}^{r-1}\binom{r}{s}\left(\frac{1}{2}\right)^{r+1}\frac{s+1-r}{r+1}}{2-\sum\limits_{s=0}^{\infty} \frac{c^s}{s!}  e^{-c} \cdot \frac{s}{r+s+1} - \left(\frac{1}{2}\right)^{r+1}\frac{r}{r+1} - \sum\limits_{s=0}^r \binom{r}{s}  \left(\frac{1}{2}\right)^{r+1} \cdot \frac{1}{s+1}}
\end{split}
\end{equation*}

Now computing the population drift is exactly the same exercise as for the \toea, except that we have to account for crossovers in the initial states. We only have to adjust the probability $\mathbf{P}[\mathcal{E}_{\text{progress}}]$ and beware that mutations now have an additional $\frac{1}{2}$ factor. Then we can write down the expression for the conditional drift of the \toga.

\begin{equation*}
\begin{split}
&\mathbf{P}[\mathcal{E}_{\text{progress}}] = \frac{1}{2} - \frac{1}{2} \left(\mathbb{P}[\mathcal{E}_0^0 \land \mathcal{E}_1^0] + \mathbb{P}[\mathcal{E}_0^1 \land \neg \mathcal{E}_1^0 \land \neg \mathcal{E}_{acc}]\right)\\\\
&\E[X_i - X_{i+1}\mid X_{i} = y \land \mathcal{E}_{\text{progress}}] = \frac{\mathbb{P}[\mathcal{E}_0^1 \land \mathcal{E}_1^0] + \sum_{r = 1}^{r_{max}}\mathbb{P}[\mathcal{E}_0^1 \land \mathcal{E}_1^r \land \mathcal{E}_{acc}] \cdot \overline{F}(r)}{2\mathbb{P}[\mathcal{E}_{\text{progress}}]}
\end{split}
\end{equation*}

\tikzstyle{level 2}=[level distance=5cm, sibling distance=6cm]
\tikzstyle{level 4}=[level distance=3cm, sibling distance=5cm]

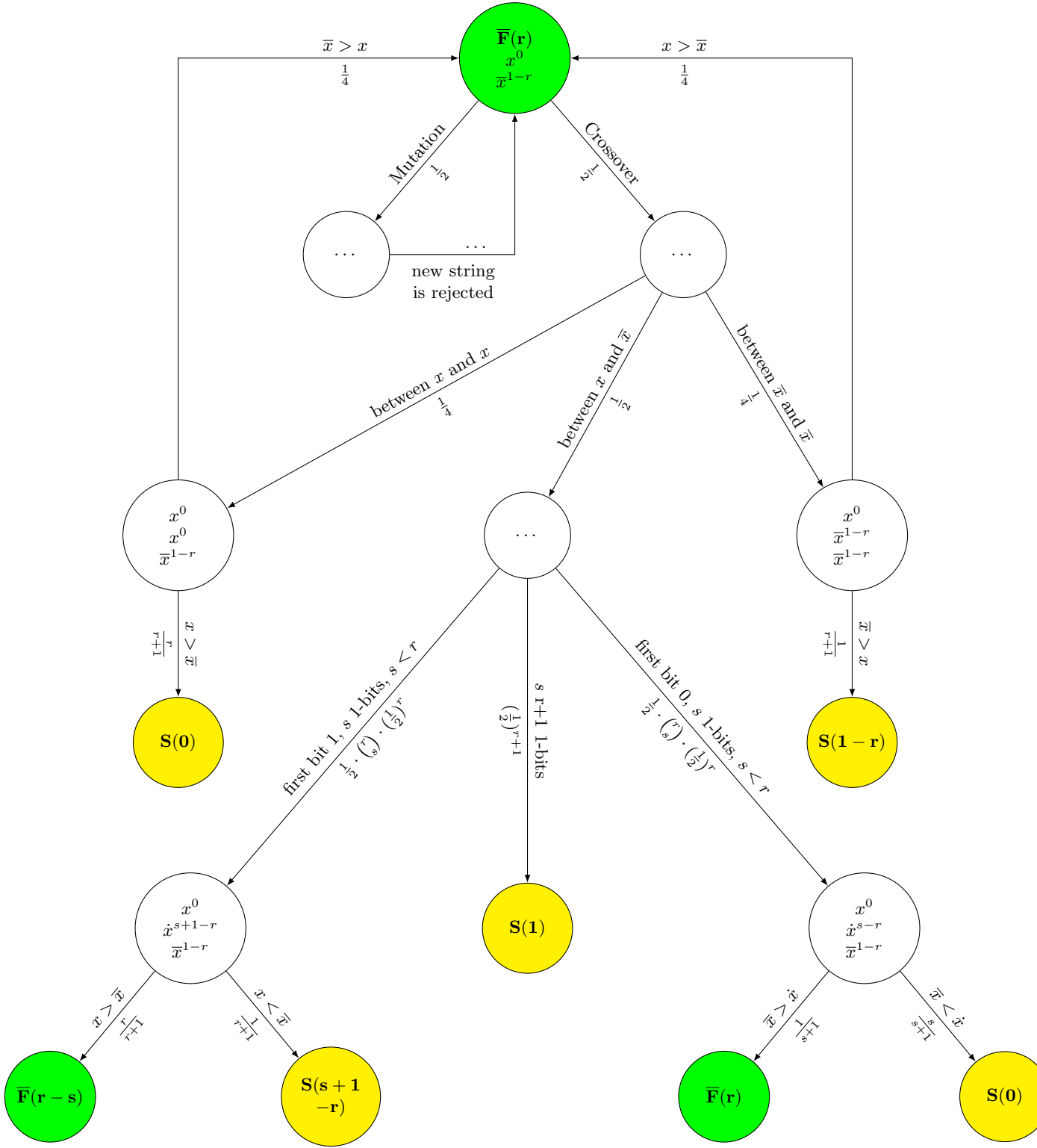
\begin{figure}[H]
\centering
\hspace*{-5em}
\begin{tikzpicture}[sloped,edge from parent/.style={draw,-latex}]
\node[bag, draw, circle, fill = green] (Origin) {$\mathbf{\overline{F}(r)}$ \\ $x^{0}$ \\ $\overline{x}^{1-r}$}
    child{
        node[bag, draw, circle] (Mutation) {$\dots$}
    edge from parent
        node[above] {Mutation}
        node[below] {$\frac{1}{2}$}
    }
    child{
        node[bag, draw, circle] (Crossover) {$\dots$}
        child{
            node[bag, draw, circle,left = 2cm] (xx) {$x^0$ \\ $x^0$ \\ $\overline{x}^{1-r}$}
            child{
                node[bag, draw, circle,fill = yellow, above = 2.5cm] {$\mathbf{S(0)}$}
            edge from parent
                node[above] {$x > \overline{x}$}
                node[below] {$\frac{r}{r+1}$}
            }
        edge from parent
            node[above] {between $x$ and $x$}
            node[below] {$\frac{1}{4}$}
        }
        child{
            node[bag, draw, circle,left = 2cm] {$\dots$}
            child{
                node[bag, draw, circle] {$x^0$ \\ $\dot{x}^{s+1-r}$ \\ $\overline{x}^{1-r}$}
                child{
                    node[bag, draw, circle, fill = green] {$\mathbf{\overline{F}(r-s)}$}
                edge from parent
                    node[above] {$x > \overline{x}$}
                    node[below] {$\frac{r}{r+1}$}
                }
                child{
                    node[bag, draw, circle, fill = yellow] {$\mathbf{S(s+1}$ $\mathbf{-r)}$}
                edge from parent
                    node[above] {$x < \overline{x}$}
                    node[below] {$\frac{1}{r+1}$}
                }
            edge from parent
                node[above] {first bit $1$, $s$ 1-bits, $s < r$}
                node[below] {$\frac{1}{2} \cdot \binom{r}{s} \cdot (\frac{1}{2})^r$}
            }
            child{
                node[bag, draw, circle, fill = yellow] {$\mathbf{S(1)}$}
            edge from parent
                node[above] {$s$ r+1 1-bits}
                node[below] {$(\frac{1}{2})^{r+1}$}
            }
            child{
                node[bag, draw, circle] {$x^0$ \\ $\dot{x}^{s-r}$ \\ $\overline{x}^{1-r}$}
                child{
                    node[bag, draw, circle, fill = green] {$\mathbf{\overline{F}(r)}$}
                edge from parent
                    node[above] {$\overline{x} > \dot{x}$}
                    node[below] {$\frac{1}{s+1}$}
                }
                child{
                    node[bag, draw, circle, fill = yellow] {$\mathbf{S(0)}$}
                edge from parent
                    node[above] {$\overline{x} < \dot{x}$}
                    node[below] {$\frac{s}{s+1}$}
                }
            edge from parent
                node[above] {first bit $0$, $s$ 1-bits, $s < r$}
                node[below] {$\frac{1}{2} \cdot \binom{r}{s} \cdot (\frac{1}{2})^r$}
            }
        edge from parent
            node[above] {between $x$ and $\overline{x}$}
            node[below] {$\frac{1}{2}$}
        }
        child{
            node[bag, draw, circle,left = 2cm] (overxx) {$x^0$ \\ $\overline{x}^{1-r}$ \\ $\overline{x}^{1-r}$}
            child{
                node[bag, draw, circle,fill = yellow, above = 2.5cm] {$\mathbf{S(1-r)}$}
            edge from parent
                node[above] {$\overline{x} > x$}
                node[below] {$\frac{1}{r+1}$}
            }
        edge from parent
            node[above] {between $\overline{x}$ and $\overline{x}$}
            node[below] {$\frac{1}{4}$}
        }
    edge from parent
        node[above] {Crossover}
        node[below] {$\frac{1}{2}$}
    };
    
\draw (xx) [out = 90, in = 180, -latex]  |- (Origin) node[above left = 0cm and 2.5cm] {$\overline{x} > x$} node[below left=0cm and 2.8cm] {$\frac{1}{4}$};
\draw (overxx) [out = 90, in = 0, -latex]  |- (Origin) node[above right = 0cm and 2.5cm] {$x > \overline{x}$} node[below right=0cm and 2.8cm] {$\frac{1}{4}$};
\draw (Mutation) node[below right = 0cm and 1cm] {\begin{tabular}{c} new string \\ is rejected \end{tabular}} node[above right=0cm and 2cm] {$\dots$} [out = 90, in = 0, -latex]  -| (Origin);

\end{tikzpicture}
\caption{Transition diagram of a degenerated population in the \toga}
\label{StateGADrift}
\end{figure}

}
{% if journal = false, no appendix
}

\bibliographystyle{splncs04}
\bibliography{refs}

\end{document}